\documentclass[letterpaper]{article} % DO NOT CHANGE THIS
\usepackage{aaai2026}  % DO NOT CHANGE THIS
\usepackage{times}  % DO NOT CHANGE THIS
\usepackage{helvet}  % DO NOT CHANGE THIS
\usepackage{courier}  % DO NOT CHANGE THIS
\usepackage[hyphens]{url}  % DO NOT CHANGE THIS
\usepackage{graphicx} % DO NOT CHANGE THIS
\usepackage{natbib}  % DO NOT CHANGE THIS AND DO NOT ADD ANY OPTIONS TO IT
\usepackage{caption} % DO NOT CHANGE THIS AND DO NOT ADD ANY OPTIONS TO IT
\usepackage{algorithm}
\usepackage{algorithmic}
\usepackage{multirow}

\usepackage{newfloat}
\usepackage{listings}
\DeclareCaptionStyle{ruled}{labelfont=normalfont,labelsep=colon,strut=off} % DO NOT CHANGE THIS
\floatstyle{ruled}
\newfloat{listing}{tb}{lst}{}
\floatname{listing}{Listing}
\title{RayD3D: Distilling Depth Knowledge Along the Ray for Robust Multi-View 3D Object Detection}
\author{
    Rui Ding\textsuperscript{\rm 1}, 
    Zhaonian Kuang\textsuperscript{\rm 1}, 
    Zongwei Zhou\textsuperscript{\rm 2}, 
    Meng Yang \textsuperscript{\rm 1}\thanks{Corresponding author.}, 
    Xinhu Zheng \textsuperscript{\rm 3}$^*$, 
    Gang Hua \textsuperscript{\rm 4}
}
\affiliations{
    \textsuperscript{\rm 1} Xi’an Jiaotong University, Xi’an, China \\
    \textsuperscript{\rm 2} Horizon Robotics, Beijing, China \\
    \textsuperscript{\rm 3} Hong Kong University of Science and Technology (Guangzhou), Guangzhou, China\\
    \textsuperscript{\rm 4} Amazon, Bellevue, United States

    \{dingrui, znkwong\}@stu.xjtu.edu.cn,  ZongweiZhou@horizon.ai, mengyang@mail.xjtu.edu.cn,\\
    xinhuzheng@hkust-gz.edu.cn, ganghua@gmail.com \\
}

\usepackage{bibentry}
\begin{document}

\maketitle

\begin{abstract}
Multi-view 3D detection with bird’s eye view (BEV) is crucial for autonomous driving and robotics, but its robustness in real-world is limited as it struggles to predict accurate depth values. A mainstream solution, cross-modal distillation, transfers depth information from LiDAR to camera models but also unintentionally transfers depth-irrelevant information (e.g. LiDAR density). To mitigate this issue, we propose RayD3D, which transfers crucial depth knowledge along the ray: a line projecting from the camera to true location of an object. It is based on the fundamental imaging principle that predicted location of this object can only vary along this ray, which is finally determined by predicted depth value. Therefore, distilling along the ray enables more effective depth information transfer. More specifically, we design two ray-based distillation modules. Ray-based Contrastive Distillation (RCD) incorporates contrastive learning into distillation by sampling along the ray to learn how LiDAR accurately locates objects. Ray-based Weighted Distillation (RWD) adaptively adjusts distillation weight based on the ray to minimize the interference of depth-irrelevant information in LiDAR. For validation, we widely apply RayD3D into three representative types of BEV-based models, including BEVDet, BEVDepth4D, and BEVFormer. Our method is trained  on clean NuScenes, and tested on both clean NuScenes and RoboBEV with a variety types of data corruptions. Our method significantly improves the robustness of all the three base models in all scenarios without increasing inference costs, and achieves the best when compared to recently released multi-view and distillation models.
\end{abstract}


% % % % % % % % % % % % % % % % % % % % % % % % % % % 
\begin{figure}[t]
    \centering
    \includegraphics[width=1.0\linewidth]{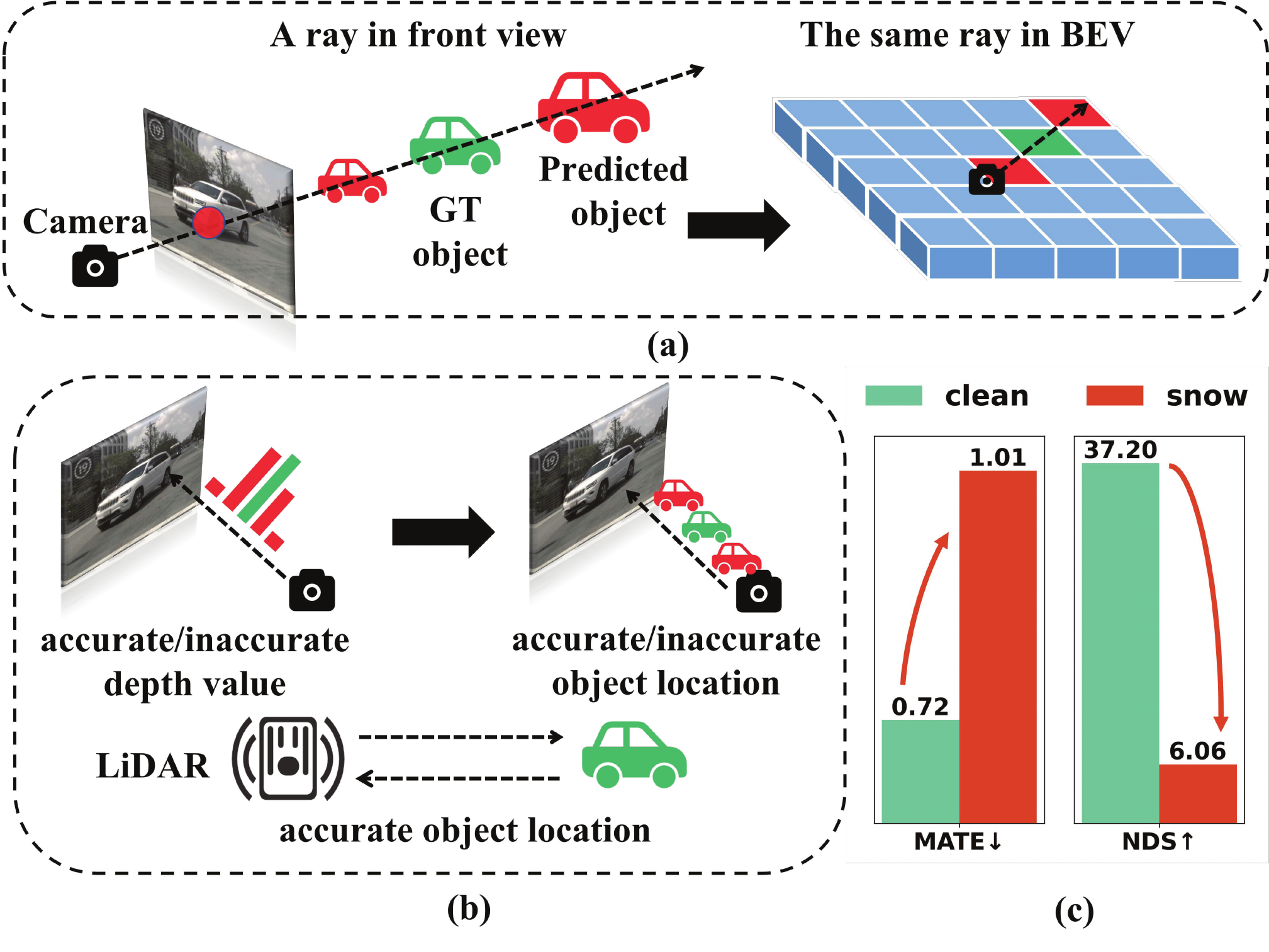}
    \caption{Ray prior for cross-modal distillation. (a) The line projecting from the camera to true location of an object forms a ray in both front view and BEV. Predicted location of this object can only vary along this ray, which is finally determined by predicted depth value. (b) LiDAR has accurate depth. However, inaccurate depth from camera leads to inaccurate location of object along the ray. Therefore, distilling along the ray enables more effective depth transfer. (c) The accuracy (MATE) of predicted depth drops from 0.72 to 1.00 when data corruptions occur, consequently, the accuracy (NDS) of 3D detection drops from 37.20 to 6.06. 
    %Ray prior in multi-view BEV detection model.
}
    \label{fig:key_idea}
\end{figure}
% % % % % % % % % % % % % % % % % % % % % % % % % % % 

% % % % % % % % % % % % % % % % % % % % % % % % % % % 
\begin{figure*}[t]
    \centering
    \includegraphics[width=0.9\linewidth]{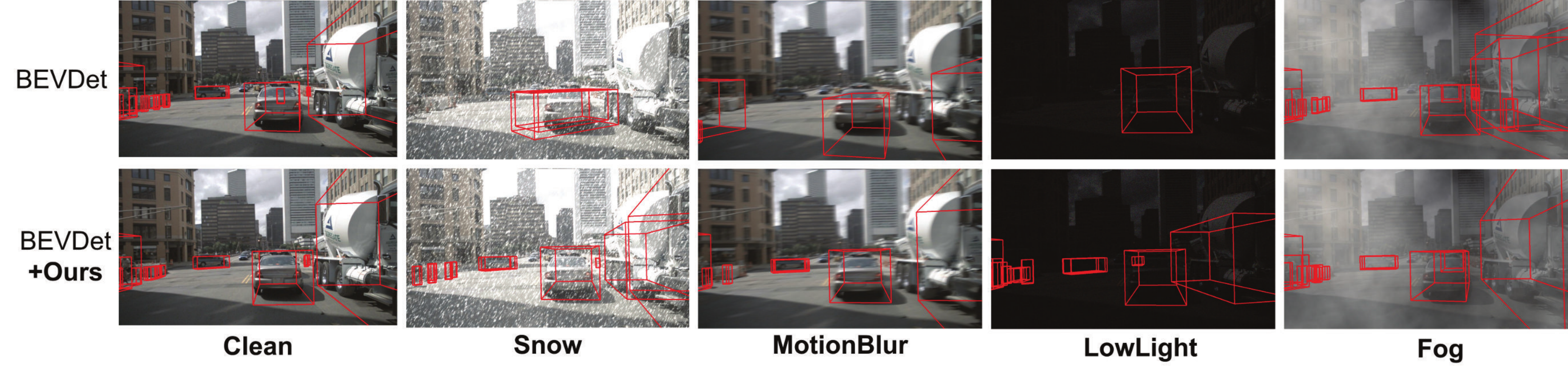}
    \caption{Visual examples of our RayD3D. The accuracy of multi-view 3D detection drops significantly in real-world scenarios. Our method consistently demonstrates strong robustness on both clean data and various types of data corruptions, whether data corruptions are seen or not during training. 
    }
    \label{fig:visual}
\end{figure*}
% % % % % % % % % % % % % % % % % % % % % % % % % % % 

\section{Introduction}

Multi-view 3D object detection with bird’s eye view (BEV) is widely applied in autonomous driving and robotics navigation. They offer a unified feature representation space for multi-view images and perform well on benchmarks with clean data. However, their performance drops significantly in real-world scenarios especially when data corruptions occur such as fog and snow. Figure \ref{fig:key_idea}(c) show that a key bottleneck is that they struggle to predict accurate depth values and data corruptions further worse the problem. 

Since LiDAR provides precise depth information \cite{zhou2018voxelnet, sun2020towards}, distilling LiDAR features to the camera offers a promising solution to mitigate this problem. However, existing cross-modal distillation methods \cite{gong2022cmkd,ding2024MonoSTL} simply force camera features to mimic LiDAR features or require a teacher network with the same structure as the student \cite{huang2024VCD,wu2023ADD}. In these methods, LiDAR models unintentionally transfer depth-irrelevant information such as point cloud density and reflection intensity, to camera models. Therefore, these methods fail to effectively transfer crucial depth information to camera models.

In this paper, we apply the ray prior in Figure \ref{fig:key_idea} to more effectively transfer depth information from LiDAR to camera. The ray refers to the line projecting from the camera to true location of an object during imaging. In multi-view 3D detection, predicted location of this object can only vary along this ray in both front view and BEV view as shown in Figure \ref{fig:key_idea}(a). 
The location of this object is finally determined by predicted depth value, as shown in Figure \ref{fig:key_idea}(b). Inaccurate depth values lead to inaccurate locations of the object along the ray, thereby, it reduces the detection accuracy. 
Therefore, distilling along the ray enables more effective depth information transfer from LiDAR to camera model.

The ray prior is a fundamental principle during imaging, which can be used in different ways in relevant 3D tasks such as NeRF \cite{2020NeRF} and pose estimation \cite{2022Ray3D}. The recent RayDN \cite{liu2024ray} uses rays in multi-view models to handle redundant predictions. However, they applied the ray in single-modal images. 

In this paper, we propose a novel method, namely RayD3D, to more effectively distill depth knowledge along the ray for robust 3D object detection in real-world scenarios. We unify LiDAR and camera features in the BEV space and successfully apply the ray prior into cross-modal distillation. It can be combined with previous ray-based multi-view methods to achieve better performance. Our method includes two ray-based distillation modules. (I) \textbf{Ray-based Contrastive Distillation} (\textbf{RCD}) module incorporates contrastive learning into distillation by sampling along the ray to learn how LiDAR accurately locates objects. (II) \textbf{Ray-based Weighted Distillation} (\textbf{RWD}) module adaptively adjusts distillation weights based on rays to minimize interference from depth-irrelevant information in LiDAR.

The RCD module incorporates contrastive learning \cite{sun2023calico} by constructing positive and negative sample pairs along the ray. Camera BEV features with accurate depth are sampled as positives, while other BEV features along the ray are negatives. By contrasting them with LiDAR features, the camera model is enhanced to learn LiDAR's knowledge to distinguish accurate and inaccurate object locations along rays. Contrastive learning ensures that the model focuses on learning depth knowledge from LiDAR rather than simply mimicking LiDAR features.

The RWD module adaptively adjusts distillation weights based on rays to control the amount of transferred LiDAR information. For rays with large differences between camera and LiDAR features, we increase weight to transfer more depth information for correction. For rays with small differences, we reduce weight to avoid interfering camera model’s own localization ability. The adaptive weight balances the benefits of transferred depth information and the interference from depth-irrelevant information.

 Our RayD3D is compatible with all dominant types of BEV detection models, including LSS-based \cite{philion2020LSS}, Transformer-based \cite{li2022bevformer}, and temporal models \cite{huang2022bevdet4d}. For validation, we widely apply RayD3D into three representative models: BEVDet \cite{huang2021bevdet}, BEVFormer \cite{li2022bevformer}, and BEVDepth4D \cite{li2023bevdepth}. Our method is trained on nuScenes with clean data, and evaluated on both nuScenes and RoboBEV with a variety types of data corruptions. 
 
 Extensive experiments show that our RayD3D significantly improves the accuracy and robustness of the three base models in all scenarios such as the examples in Figure \ref{fig:visual}. The performance of our method further improves when trained on corrupted data. Our method also achieves the best performance when compared to recently released multi-view and distillation models with the same setting.

Our main contributions are highlighted as follows.

\begin{itemize}
    \item We are the first to apply the ray prior for cross-modal knowledge distillation, which enables more effective depth information transfer from LiDAR to camera.
    \item We propose two novel ray-based distillation modules to enhance the robustness of 3D detection. Our method is compatible with dominant types of BEV-based models.
    \item Our method significantly boosts the robustness of three representative base models on both clean nuScenes and RoboBEV with a variety types of data corruptions.

\end{itemize}

% % % % % % % % % % % % % % % % % % % % % % % % % % % 
\begin{figure*}[t]
    \centering
    \includegraphics[width=1.0\linewidth]{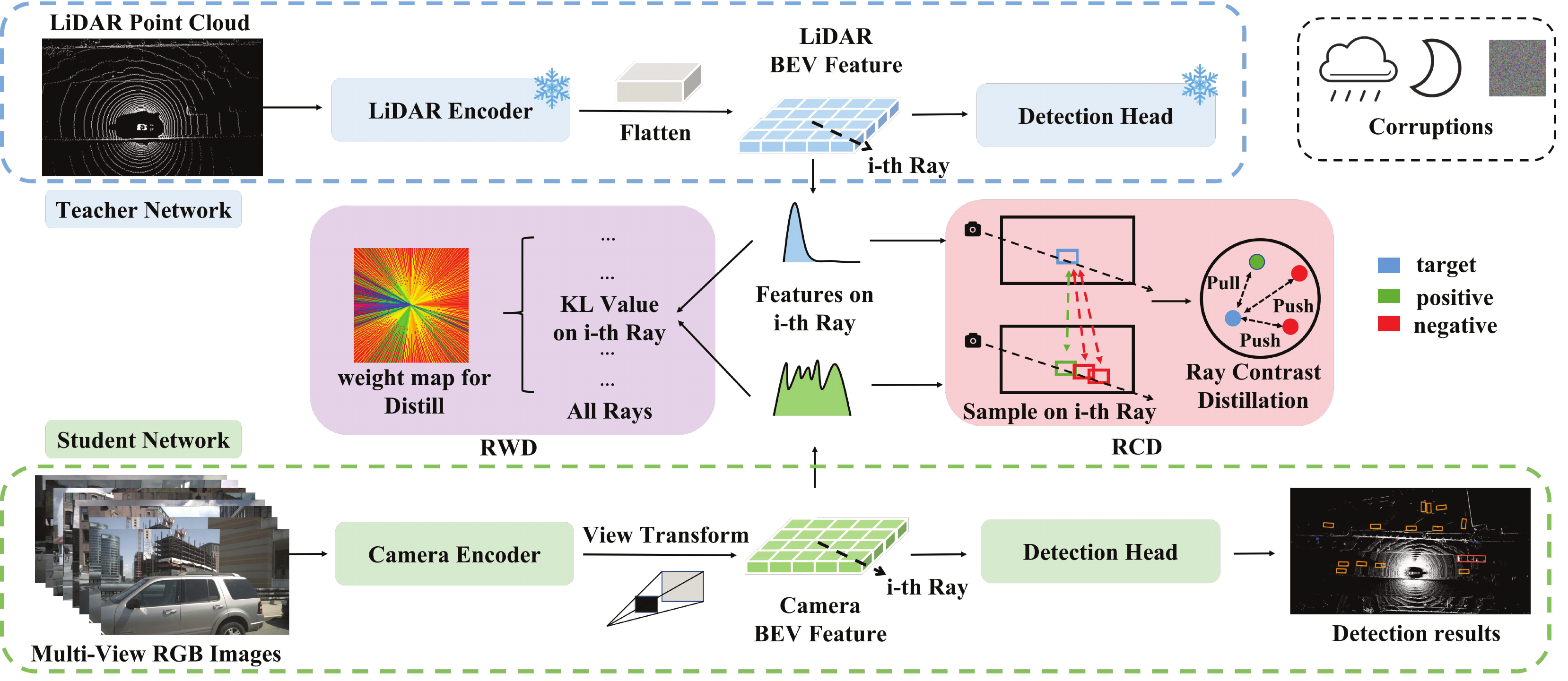}
    \caption{Overview of our RayD3D framework. It includes a teacher network, a student network, and two novel ray-based distillation modules. First, we train the LiDAR-based teacher network and freeze its parameters. The student network is then initialized randomly and trained with our \textbf{RCD} and \textbf{RWD} distillation modules. During inference, only the student network is used and evaluated on both clean and corrupted data. It significantly enhances the accuracy and robustness in real-world scenarios without increasing inference costs.
}
    \label{fig:framework}
\end{figure*}
% % % % % % % % % % % % % % % % % % % % % % % % % % % 

\section{Related Work}
\subsection{Multi-View 3D Detection}
Multi-View 3D object detection \cite{wang2021fcos3d,wang2022detr3d} predicts objects' dimensions, locations, and orientations from images. Recently, BEV-based models have become dominant and the core is transforming images into BEV. Thus, these models can be divided into two categories: (I) LSS-based (Lift, Splat, Shoot) \cite{philion2020LSS,li2023bevdepth}. They lift camera features to 3D space and project them onto the BEV. However, inaccurate depth estimation affects detection accuracy. (II) Transformer-Based \cite{li2022bevformer}. They use deformable attention to dynamically sample features from images and map them to the BEV. This allows a single pixel to map to multiple BEV locations, which can also result in inaccurate locations. In addition, models like BEVDepth4D \cite{huang2022bevdet4d}, incorporate temporal information and achieve detection accuracy close to LiDAR-based detectors.

\subsection{Data Corruptions}

However, most existing multi-view models cannot well handle data corruptions \cite{xie2023robobev,kong2024robodepth} in real world, such as fog. To enhance robustness, researchers uses denoising methods \cite{guo2020low-light} or simulate severe weather \cite{qian2024weatherdg,gupta2024robust_weather}. These strategies help models perform well on specific data corruptions, however, may fail on other types of data corruptions in real world. It is impractical to denoise or simulate all types of unexpected data corruptions in real world.

\subsection{Cross-Modal Knowledge Distillation}

Knowledge distillation was originally proposed for model compression \cite{hinton2015distilling} and then widely applied in various domains \cite{yang2022cross,chen2017learning}. In multi-view 3d detection tasks, cross-modal knowledge distillation methods \cite{klingner2023x3kd,zhao2024simdistill} like BEVDistill transfer information from LiDAR teacher network to camera student network. LiDAR-based teacher network can compensate for the limitation of camera model. To reduce modality gap, recent methods \cite{kim2024labeldistill,zhou2023unidistill,huang2024VCD,wu2023ADD} transform features onto the BEV and train student BEV features to mimic those of the teacher. However, these methods fail to effectively transfer crucial depth information and do not fully consider camera model robustness in real world. In contrast, we successfully apply the ray prior into cross-modal distillation to improve the robustness of camera models.

\section{Method}
\label{sec: Method}

\subsection{Overall Framework}

Figure \ref{fig:framework} shows our framework, which is designed to improve performance of the multi-view models in all scenarios. It includes three main components: (I) The teacher network takes LiDAR input to generate BEV features; (II) the student network takes $N_{camera}$ camera images as input and transforms features into BEV perspective; (III) two novel ray-based distillation modules, Ray-based Contrastive (RCD) and Ray-based Weighted (RWD) transfer crucial depth information to enhance the camera model's performance.

More specifically, during distillation process, we use two convolutional layers to resize feature maps of camera to the same size as LiDAR feature maps. BEV feature maps of the camera $C\in R^{D\times H \times W }$ and BEV feature maps of the LiDAR $L\in R^{D\times H \times W }$ are used for distillation, where $D$, $H$, and $W$ represent the dimensions of the BEV feature maps. For distilling along the rays, we evenly divide the BEV feature map into $N_{ray}$ rays.

\subsection{Ray-Based Contrastive Distillation (RCD)}
\label{subsec:RCD}

The RCD module guides the model to learn how LiDAR distinguishes accurate and inaccurate locations along rays. By leveraging the ray prior during the front view to BEV transformation \cite{zhang2023DAbev,liu2024ray,chu2024rayformer}, it transfers depth information more effectively than directly mimicking LiDAR features. This is achieved by sampling positive and negative pairs along the rays, as shown in Figure \ref{fig:framework}.

For positive samples on the $i$-th ray, we sample features from the object's foreground region, with the camera and LiDAR features at the same $j$-th position denoted as $C_{i,j}$ and $L_{i,j}$, respectively. It ensures that the sampled positive features correspond to the accurate object locations. 

For negative samples, we prioritize sampling from the inaccurate locations near positive samples, ensuring that they are similar in camera features but differ significantly in depth location. This enhances the effectiveness of distinguishing accurate and inaccurate locations. Specifically, We adopt a Gaussian-based sampling strategy to control the probability of negative samples, making the position closer to the positive sample more likely to be selected. 

The negative sample from the $k$-th position on the $i$-th ray is denoted as $C_{i,k}$. The sampling probability $P(C_{i,k})$:

% % % % % % % % % % % % % % % % % % % % % % % % % % % 
\begin{equation}
  P( C_{i,k}) = \frac{\exp\left(-\frac{d(i,j,k)^2}{2\sigma^2}\right)}{\sum_{l} \exp\left(-\frac{d(i,j,l)^2}{2\sigma^2}\right)} \quad ,
  \label{eq:neg_sample}
\end{equation}
% % % % % % % % % % % % % % % % % % % % % % % % % % % 
where $d(i,j,k)$ represents the distance between samples $C_{i,j}$ and $C_{i,k}$. Smaller $\sigma$ values increase the sampling probability of position closer to the positive sample. Each ray samples one positive sample and $N_{neg}$ negative samples. Notably, instead of selecting individual pixel features from the BEV feature map, we sample from an $m \times m$ region around each location to capture local context. This helps improve the stability of the loss during distillation.

After constructing the positive and negative pairs, we compute the ray-based contrastive distillation loss $L_{RCDS}$ on the student network and and $L_{RCDT}$ on the teacher network, respectively. They are defined as follows:

% % % % % % % % % % % % % % % % % % % % % % % % % % % 
\begin{equation}
L_{RCDS}= {\textstyle \sum_{i=1}^{N_{ray}}} \frac{exp(L_{i,j} \cdot C_{i,j}/\tau  ) }{ {\textstyle \sum_{k=1}^{N_{neg}+1}}exp(L_{i,j} \cdot C_{i,k}/\tau ) + \xi  }\ ,
\label{eq:loss_RCDS}
\end{equation}
% % % % % % % % % % % % % % % % % % % % % % % % % % % 

% % % % % % % % % % % % % % % % % % % % % % % % % % % 
\begin{equation}
L_{RCDT}= {\textstyle \sum_{i=1}^{N_{ray}}} \frac{exp(C_{i,j} \cdot L_{i,j}/\tau  ) }{ {\textstyle \sum_{k=1}^{N_{neg}+1}}exp(C_{i,j} \cdot L_{i,k}/\tau ) + \xi  }\ , 
\label{eq:loss_RCDT}
\end{equation}
% % % % % % % % % % % % % % % % % % % % % % % % % % % 
where $\tau$ is the temperature parameter to adjust the contrastive strength between positive and negative samples. The factor $\xi$ is set to a constant value $1e-6$ to prevent zero in the denominator. $L_{i,k}$ refers to the negative sample from the $k$-th position on the $i$-th ray of LiDAR feature maps. Computing contrastive loss \cite{he2020InfoNCELoss,tian2019contrastive} in both the student and teacher networks increases the diversity of negative examples, further improving the effectiveness of contrastive learning. Finally, the total RCD loss is defined as follows:
 
% % % % % % % % % % % % % % % % % % % % % % % % % % % 
\begin{equation}
L_{RCD}=-log(\frac{1}{N_{ray}} (L_{RCDS}+L_{RCDT})) \ . 
\label{eq:loss_RCD}
\end{equation}
% % % % % % % % % % % % % % % % % % % % % % % % % % % 

As shown in Figure \ref{fig:framework}, the loss pulls positive pair features closer while pushing negative pair features further apart, improving the model’s ability to distinguish accurate and inaccurate locations. This finally enhances the robustness.

\subsection{Ray-Based Weighted Distillation (RWD)}
\label{subsec:RWD}

The RWD module minimizes interference from depth-irrelevant information by adaptively adjusting distillation weights. The weights are set based on depth estimation accuracy of different regions. However, models like BEVFormer do not explicitly predict depth, so depth error cannot be directly computed. As shown in Figure \ref{fig:framework}, LiDAR has precise depth information, with high activation of features only at foreground locations along a ray. Camera has inaccurate depth estimation, causing high activation in multiple areas. If the feature distribution of the camera is similar to LiDAR along the ray, it means the foreground object's location in the camera model is precise. 

By analyzing feature differences along each ray, we can determine the accuracy of depth estimation and the amount of depth information needed by camera. To measure the difference between Camera and LiDAR features in each ray, we use Kullback-Leibler (KL) divergence. However, directly calculating KL divergence on the original feature maps is challenging. Feature maps with too many channels make the distillation less stable and difficult to optimize. Therefore, we first generate spatial attention maps for camera and LiDAR, denoted as $SC$ and $SL$, respectively:

% % % % % % % % % % % % % % % % % % % % % % % % % % % 
\begin{equation}
SC=softmax(\frac{1}{D} {\textstyle \sum_{d=1}^{D}\left | C^d \right | }  ) \ , 
\end{equation}
% % % % % % % % % % % % % % % % % % % % % % % % % % % 
% % % % % % % % % % % % % % % % % % % % % % % % % % % 
\begin{equation}
SL=softmax(\frac{1}{D} {\textstyle \sum_{d=1}^{D}\left | L^d \right | }  ) \ , 
\end{equation}
% % % % % % % % % % % % % % % % % % % % % % % % % % % 
where $C^d$ and $L^d$ are the $d$-th channel of camera and LiDAR features, respectively. $SC \in R^{H \times W}$ and $SL \in R^{H \times W}$ are the spatial attention maps for camera and LiDAR, respectively. This strategy reduces the impact of the modality gap between camera and LiDAR.

% % % % % % % % % % % % % % % % % % % % % % % % % % % 
\begin{figure}[t]
    \centering
    \includegraphics[width=1.0\linewidth]{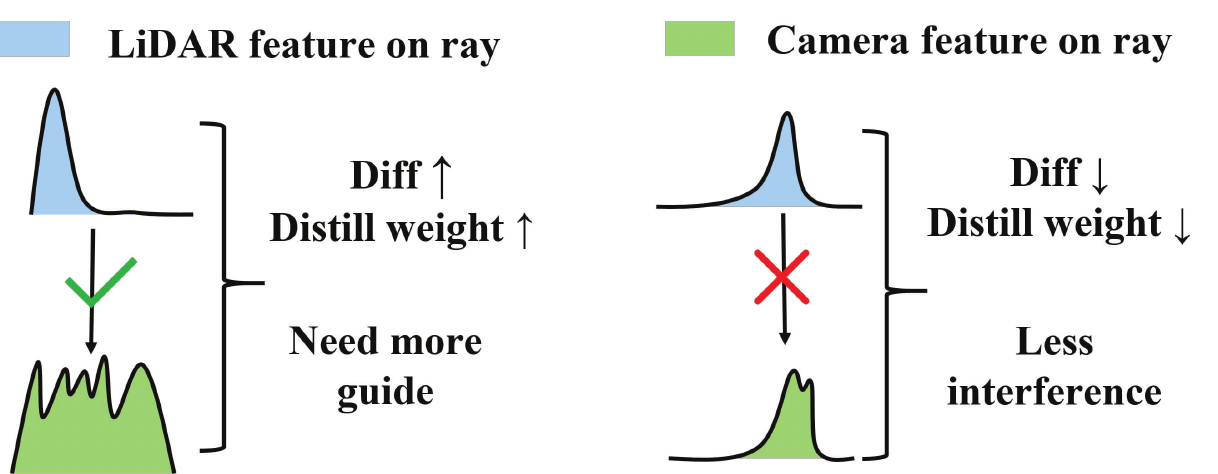}
    \caption{Distillation weight in RWD. For rays with large differences between camera and LiDAR features, we increase weight to transfer more depth information for correction. For rays with small differences, we reduce weight to avoid the interference of depth-irrelevant information on the camera model itself.
}
    \label{fig:RWD}
\end{figure}
% % % % % % % % % % % % % % % % % % % % % % % % % % % 

Next, we compute feature difference along the $i$-th ray: $A_i = KL(SL_i, SC_i)$. $SL_i$ and $SC_i$ represent the LiDAR and camera features along the $i$-th ray. As shown in Figure \ref{fig:RWD}, a large $A_i$ indicates the camera model predicts depth inaccurately, so we increase the distillation weight to transfer more LiDAR depth information for correction. A small $A_i$ means the camera model predicts depth accurately on its own. We reduce the weight to limit interference from depth-irrelevant information. Then, we apply each ray's weight $A_i$ to all the points along its ray in BEV as shown in Figure \ref{fig:framework}. The generated weight map $\omega $ is as follows:

% % % % % % % % % % % % % % % % % % % % % % % % % % % 
\small
\begin{equation}
\omega [h,w]= \frac{1}{I((h,w) \in R_i)}{\textstyle \sum_{i=1}^{N_{ray}} A_i\cdot I((h,w) \in R_i)  }  \ , 
\label{eq:loss_weight}
\end{equation}
\normalsize
% % % % % % % % % % % % % % % % % % % % % % % % % % % 
where $\omega \in R^{H \times W}$ represents final weight map, and $R_i$ denotes all coordinates along the $i$-th ray. The indicator function $I((h,w) \in R_i)$ determines whether coordinate $[h, w]$ lies on ray $R_i$. Inspired by previous works \cite{wang2023distillbev,chongmonodistill}, we scale the background weights by a constant factor $s$ to reduce its influence in distillation. 

% we also apply different weights to foreground and background areas. After calculating the weights along each ray

To ensure consistent weighting within each foreground object, all regions inside the same object are assigned the same weight. Specifically, this weight is determined by taking the maximum value among all rays passing through the object. Finally, the ray-based weighted distillation loss $L_{RWD}$ is as follows:

% % % % % % % % % % % % % % % % % % % % % % % % % % % 
\small
\begin{equation}
L_{RWD}=\frac{1}{H\times W} { \sum_{h=1}^{H}}  {\sum_{w=1}^{W}} (\omega[h,w] \cdot | L[h,w]-C[h,w] | ) \ . 
\label{eq:loss_RWD}
\end{equation}
\normalsize
% % % % % % % % % % % % % % % % % % % % % % % % % % % 

In total, the adaptive weighting strategy enables the camera model to learn reliable depth information from LiDAR while minimizing interference. It significantly enhances depth estimation ability of the camera model, reducing the impact of corruptions on detection accuracy. In addition, it avoids overfitting to LiDAR features.

\subsection{Total Loss}

In addition to RCD and RWD, we also incorporate the classic response distillation loss \cite{ding2024MonoSTL}. Response distillation allows the student to be close to the teacher's output. The loss $L_{RES}$ is as follows:

% % % % % % % % % % % % % % % % % % % % % % % % % % % 
\begin{equation}
L_{RES}= \frac{1}{N_{heads}} \sum_{n=1}^{N_{heads}}(Y_{L,n}-Y_{C,n})  \ ,
\end{equation}
% % % % % % % % % % % % % % % % % % % % % % % % % % % 
where $N_{heads}$ denotes the number of prediction heads, and $Y_{L,n}$ and $Y_{C,n}$ represent the features of the $n$-th head for LiDAR and camera, respectively. We also use the original loss $L_{SRC}$ for 3D detection in the student network. Our total loss function is as follows:

% % % % % % % % % % % % % % % % % % % % % % % % % % % 
\begin{equation}
L=L_{RCD}+L_{RWD}+L_{SRC}+L_{RES} \ . 
\end{equation}
% % % % % % % % % % % % % % % % % % % % % % % % % % % 

This loss can be applied to all types of BEV models in the literature, enhancing their robustness in real-world scenarios without increasing inference costs.

% % % % % % % % % % % % % % % % % % % % % % % % % % % 
\begin{table*}[t]
    \centering

    \setlength{\tabcolsep}{1.5mm}

    \begin{tabular}{c|cccccccc|c|cc}
    \hline
    \multirow{3}{*}{Method} & \multicolumn{9}{c|}{percentage ($NDS_i/NDS_{clean}$) \% (RoboBEV) } & \multicolumn{2}{c}{NDS/mAP (nuScenes)} \\ \cline{2-12} 
    & Bright & Camera & Color & \multirow{2}{*}{Fog} & Frame & Low & Motion & \multirow{2}{*}{Snow} & \multirow{2}{*}{mRR} & \multirow{2}{*}{Val} & \multirow{2}{*}{Test} \\
    & ness & Crash & Quant & & Lost & Light & Blur & & & & \\ \hline
    BEVDet  & 67.9 & 66.3 & 65.4 & 64.9 & 52.5 & 29.2 & 55.1 & 16.3 & 52.2 & 37.4/29.6 & 38.4/28.9 \\

    \textbf{+ours} & \textbf{69.6} & \textbf{66.3} & \textbf{66.0} & \textbf{66.4} & \textbf{57.0} & \textbf{29.5} & \textbf{59.2} & \textbf{23.8} & \textbf{54.7} & \textbf{41.9/32.7} & \textbf{43.1/33.7} \\

    \hline 
    \hline
    BEVDepth4D & 73.3 & 68.2 & 71.3 & 76.6 & 71.0 & 41.8 & 64.8 & 22.4 & 61.1 & 48.4/36.4 & 49.0/38.2 \\

    \textbf{+ours} & \textbf{76.7} & \textbf{68.6} & \textbf{72.8} & \textbf{79.0} & \textbf{73.3} & \textbf{49.3} & \textbf{66.7} & \textbf{30.4} &  \textbf{64.6} & \textbf{50.7/39.3} & \textbf{52.0/40.3} \\

    \hline 
    \hline
    BEVFormer & 89.5 & 67.8 & 76.9 & 86.2 & 53.4 & 27.9 & \textbf{61.6} & 33.5 & 62.1 & 35.4/25.2 & - \\

    \textbf{+ours} & \textbf{91.9} & \textbf{67.9} & \textbf{78.5} & \textbf{87.0} & \textbf{59.2} & \textbf{32.2} & 59.8 & \textbf{42.0} &  \textbf{64.8} & \textbf{41.4/31.5} & - \\
    \hline 
    \hline
    BEVDepth4D-r101 & 67.8 & 69.9 & 61.4 & 79.9 & 73.5 & 19.3 & 57.1 & 18.5 & 55.9 & 53.5/41.2 & 55.0/43.3 \\

    \textbf{+ours} & \textbf{72.1} & \textbf{70.5} & \textbf{62.4} & \textbf{80.0} & \textbf{71.6} & \textbf{27.3} & \textbf{61.1} & \textbf{22.1} &  \textbf{58.4} & \textbf{55.3/45.4} & \textbf{57.6/47.3} \\
    \hline
    \end{tabular}
    \caption{Comparison to base models on nuScenes and RoboBEV. Our method significantly improves the accuracy (\textbf{NDS/mAP}) of base models on clean data and enhances the robustness (\textbf{mRR}) of all eight types of data corruptions. We use the same setting FP16 with official codes of BEVDet and BEVDepth4D to ensure a fair comparison with recent cross-modal distillation methods  DistillBEV \cite{wang2023distillbev} and VexKD \cite{yuzhe2025vexkd}.}
    \label{table:compare to base models}
\end{table*}
% % % % % % % % % % % % % % % % % % % % % % % % % % % 

% % % % % % % % % % % % % % % % % % % % % % % % % % % 
\begin{table*}[t]
    \centering

    \setlength{\tabcolsep}{1.4mm}

    \begin{tabular}{c|c|cccc|cc}
    \hline
    Method & Type & Teacher & Student & Backbone & Frame & NDS/mAP & mRR \\ \hline
    PETR* \cite{liu2022petr} & Multi-View & - & - & ResNet-50 & 1 & 36.7/31.7 & 61.3 \\
    
    BEVerse* \cite{zhang2022beverse} & Multi-View & - & - & SwinT & 1+3 & 46.6/32.1 & 48.6 \\
    Sparse4D* \cite{lin2022sparse4d} & Multi-View & - & - & ResNet-101 & 1+3 & 54.4/44.0 & 55.0 \\
    SoloFusion* \cite{park2022SoloFusion} & Multi-View & - & - & ResNet-50 & 1+16 & 48.5/38.7 & 64.4 \\
    DualBEV \cite{li2024dualbev} & Multi-View & - & - & ResNet-50 & 1+1 & 50.4/38.0 & - \\
    \hline
    Unidistill \cite{zhou2023unidistill} &  Distill & L+C & BEVDet & ResNet-50 &1 & 39.3/29.6 & 50.6\\
    DistillBEV \cite{wang2023distillbev} & Distill & L & BEVDet & ResNet-50 &1 & 40.6/31.4 & 52.4 \\
    VexKD \cite{yuzhe2025vexkd} &  Distill & L+C & BEVDet & ResNet-50 &1 & 40.6/\textbf{34.7} & - \\
    % \rowcolor{gray!20}
    \textbf{Ours} & Distill & L & BEVDet & ResNet-50 &1 & \textbf{41.9}/32.7 & \textbf{54.7} \\
    \hline
    \hline
    TiG-BEV \cite{huang2022tig} &  Distill & L & BEVDepth4D & ResNet-50 &1+1  & 46.1/36.6 & 62.2 \\
    DistillBEV \cite{wang2023distillbev} & Distill & L & BEVDepth4D & ResNet-50 &1+1 & 50.3/38.3 & 62.6 \\
    LabelDistill \cite{kim2024labeldistill} & Distill & L & BEVDepth4D & ResNet-50 &1+1 & 50.1/37.9 & - \\
    % \rowcolor{gray!20}
    \textbf{Ours} & Distill & L & BEVDepth4D & ResNet-50 &1+1 & \textbf{50.7/39.3} & \textbf{64.6} \\
    \hline
    \hline
    DistillBEV \cite{wang2023distillbev} & Distill & L & BEVDepth4D & ResNet-101 &1+1 & 53.1/42.7 & 58.4 \\
    % \rowcolor{gray!20}
    \textbf{Ours} & Distill & L & BEVDepth4D & ResNet-101 &1+1 & \textbf{54.8/44.0} & \textbf{59.4} \\
    \hline
    \end{tabular}
    \caption{Comparison to recent models. ‘L+C’ and ‘L’ indicate that the teacher network uses a fusion model (LiDAR + Camera) and a single LiDAR model, respectively. * refers to the results are obtained from RoboBEV official github \cite{xie2023robobev}. ‘1+1’ indicates using current and one previous frame. For fair comparison, we use the version of LabelDistill with 2 frames.}
    \label{table::Comparison other models}
\end{table*}
% % % % % % % % % % % % % % % % % % % % % % % % % % % 

% % % % % % % % % % % % % % % % % % % % % % % % % % % 
\begin{table*}[t]
    \centering

    \setlength{\tabcolsep}{1.6mm}

    \begin{tabular}{c|cccccccc|c|c}
    \hline
    \multirow{3}{*}{Method} & \multicolumn{9}{c|}{percentage ($NDS_i/NDS_{clean}$) \% (RoboBEV)} & NDS/mAP (nuScenes)\\ \cline{2-11} 
    & Bright & Camera & Color & \multirow{2}{*}{Fog} & Frame & Low & Motion & \multirow{2}{*}{Snow} & \multirow{2}{*}{mRR} & \multirow{2}{*}{Clean} \\
    & ness & Crash & Quant & & Lost & Light & Blur & & & \\ \hline
    BEVDet  & 91.7 & 69.2 & 91.6 & 93.7 & 53.0 & 55.2 & 71.1 & 81.2 & 75.8 & 38.6/29.9  \\
    +DistillBEV & 93.6 & 70.0 & 92.6 & 94.6 & 58.4 & 53.6 & 72.4 & 85.1 & 77.5 & 40.9/31.7  \\
    % \rowcolor{gray!20}
    \textbf{+ours} & \textbf{96.3} & \textbf{70.0} & \textbf{94.8} & \textbf{95.8} & \textbf{60.1} & \textbf{61.8} & \textbf{78.0} & \textbf{90.7} & \textbf{80.9} & \textbf{41.6/32.0}  \\
    \hline
    \hline
    BevDepth4D & 93.5 & \textbf{72.3} & 92.6 & 95.0 & 85.4 & 64.2 & 78.9 & 83.6 & 83.2 & 49.1/37.1  \\
    +DistillBEV & 94.8 & 71.6 & 94.4 & 96.2 & 85.8 & 67.3 & 79.1 & 88.0 & 84.7 & 50.0/38.6  \\
    % \rowcolor{gray!20}
    \textbf{+ours} & \textbf{96.1} & 72.1 & \textbf{95.6} & \textbf{96.8} & \textbf{85.8} & \textbf{69.4} & \textbf{81.8} & \textbf{92.6} & \textbf{86.3} & \textbf{50.8/38.9}  \\
    
    \hline
    \end{tabular}
    \caption{Train and test with data corruptions. The results further verify the robustness of our method, either data corruptions are seen or not during training.}
    \label{table:compare to known corruptions}
\end{table*}
% % % % % % % % % % % % % % % % % % % % % % % % % % % 

% % % % % % % % % % % % % % % % % % % % % % % % % % % 
\begin{table}[t]
    \centering

    \setlength{\tabcolsep}{4.0mm}

    \begin{tabular}{c|cc|cc}
    \hline
    & RCD & RWD & NDS/mAP & mRR \\ \hline
    a & & & 37.4/29.6 & 52.2\\
    b & $\surd$ & & 40.6/31.8 & 53.7\\
    c & & $\surd$ & 41.7/32.0 & 54.1 \\
    % \rowcolor{gray!20}
    d & $\surd$ &  $\surd$ & \textbf{41.9/32.7} & \textbf{54.7} \\

    \hline
    
    \end{tabular}
    \caption{Ablation study on our two distillation modules. }
    \label{table::Ablation studies on distillation modules}
\end{table}
% % % % % % % % % % % % % % % % % % % % % % % % % % % 

% % % % % % % % % % % % % % % % % % % % % % % % % % % 
\begin{table}[t]
    \centering

    \setlength{\tabcolsep}{1.0mm}

    \begin{tabular}{c|c|cc}
    \hline
  Sample negative & Sample in teacher & NDS/mAP & mRR \\ \hline
\multirow{2}{*}{Random} &  & 39.4/29.9 & 53.2   \\
                  & $\surd$  & 40.0/30.1  & 53.0   \\\hline
\multirow{2}{*}{Gaussian (\textbf{ours})} &  & \textbf{40.6/31.8} & 53.0   \\
                  & $\surd$  & \textbf{40.6/31.8}  & \textbf{53.7} \\

    \hline
    
    \end{tabular}
    \caption{Ablation study on sampling strategies in RCD.}
    \label{table::Ablation studies on sample strategy in RCD}
\end{table}
% % % % % % % % % % % % % % % % % % % % % % % % % % % 

% % % % % % % % % % % % % % % % % % % % % % % % % % % 
\begin{table}[t]
    \centering

    \setlength{\tabcolsep}{5.0mm}

    \begin{tabular}{c|cc}
    \hline
    Strategies  & NDS/mAP & mRR \\ \hline
    Cosine similarity & 40.2/31.1 & 53.2 \\
    JS divergence & 41.0/31.8 &  53.4 \\
    % \rowcolor{gray!20}
    KL divergence (\textbf{ours}) & \textbf{41.7/32.0} & \textbf{54.1} \\
    \hline
    
    \end{tabular}
    \caption{Ablation studies on calculation strategies of cross-modal feature difference in RWD.}
    \label{table::Ablation studies on feature difference computation strategies in RWD}
\end{table}
% % % % % % % % % % % % % % % % % % % % % % % % % % % 

\section{Experiments}
\label{sec: Experiments}

\subsection{Setting}

\subsubsection{Dataset and Metrics}

Our method is evaluated on the large-scale autonomous driving benchmarks nuScenes \cite{caesar2020nuscenes}. nuScenes includes clean images captured with six camera views and a 32-beam LiDAR, which consists of 700 training, 150 validation, and 150 testing scenes. We use the nuScenes Detection Score (NDS) and mean Average Precision (mAP) as metrics.

RoboBEV \cite{xie2023robobev} includes eight types of data corruptions based on nuScenes including Brightness, low light, Fog, Snow, Motion Blur, Color Quant, Camera Crash, and Frame Lost.  The mean Resilience Rate (\textbf{mRR}) is employed to evaluate the average performance of \textit{N} types of data corruptions as $mRR = \frac{1}{N} ({NDS_i}/NDS_{clean})$, where $NDS_{clean}$ and $NDS_i$ refer to $NDS$ values on clean data and corrupted data, respectively. Notably, RoboBEV only adds corruptions to nuScenes validation, so there is no risk of information leakage.

\subsubsection{Network Configurations}

For validation, we select three representative base models BEVDet \cite{huang2021bevdet}, BEVFormer \cite{li2022bevformer}, and BEVDepth4D \cite{li2023bevdepth}. For the teacher network, we choose the widely used CenterPoint \cite{yin2021centerpoint} for fair comparison. Both teacher and student models transform inputs into BEV feature maps for distillation and detection.

The parameters for the teacher and student networks follow those of the original models unless otherwise specified. All base models use ResNet-50 as backbone. For BEVDet and BEVDepth4D, the image size is downsampled to 256×704, while for BEVFormer, it is set to 800×450. The BEV grid size of BEVFormer is 50×50. BEVDepth4D-r101 use backbone ResNet-101 with image size 512*1408. 

The KL temperature is 0.07, and the number of negatives is 5. We fix the random seed to 0, and the variance across multiple runs is below 0.2 NDS. With our distillation modules, the training time slightly increases when training for 24 epochs on 8×3090 GPUs, while the inference time not increase.

% In RCD, the sampling region $m*m$ is set as 3×3, and in RWD, the scaling factor $s$ for background is set as 0.25.

\subsection{Evaluation}

We train all models on clean NuScenes and test them on both NuScenes and RoboBEV. It aims to verify the accuracy on clean data, and the robustness on corrupted data in real world even though data corruptions are not seen during training. We then compare our method with recent multi-view detection and cross-modal distillation methods. In all experiments, clean LiDAR data are used to ensure accurate depth information transfer. Since both teacher and student are trained on clean nuScenes training set and evaluated on corrupted RoboBEV validation set, there is no risk of information leakage. In addition, we also show the results that trained and tested models with data corruptions, which aims to further verify the robustness of our method on corrupted data.

\subsubsection{Comparison to Base Models}

 Table \ref{table:compare to base models} shows that our method significantly improves the accuracy of the three base models on clean data and all types of data corruptions. Although BEVFormer does not explicitly predict depth, it still struggles with inaccurate localization when transforming the front view to BEV. Our method well boosts its performance in all scenarios. The reason behind is that precise depth helps maintain stable BEV projections even when corruptions occur, leading to more robust result.

\subsubsection{Comparison to Recent Multi-View Models}

Table \ref{table::Comparison other models} shows that our method achieves higher accuracy on both clean data (NDS/mAP in nuScenes) and data corruptions (mRR in RoboBEV) when compared to recent multi-view models, even though some multi-view detection models use stronger backbones and higher image resolutions. 

\subsubsection{Comparison to Recent Distillation Models}

Table \ref{table::Comparison other models} shows that with the same base model, our method outperforms all recent cross-modal distillation methods on both clean and corrupted data. Notably Unidistill and VexKD use fusion models as teachers, while our method uses only LiDAR models. Our model still achieves better performance. This is because these cross-modal methods simply force camera features to mimic LiDAR features. In contrast, our RCD and RWD modules effectively guide the camera to learn relevant LiDAR depth information.

\subsubsection{Train and Test with Data Corruptions}

Our method is robust to various data corruptions in real-world scenarios, whether they are seen or not during training. Table \ref{table:compare to base models} has verified the robustness when training on clean data and test on corrupted data. To further validate, we simulate the eight types of data corruptions during training, and test on RoboBEV with data corruptions in Table \ref{table:compare to known corruptions}. In this case, the impact of data corruptions on depth is known during training, allowing our model to more effectively learn from clean LiDAR information. Our method improves the robustness of the base model BEVDet by 5.1\%, and also outperforms the representative distillation method DistillBEV.

\subsection{Ablation Studies}

We conduct ablation studies to verify the effectiveness of each module. All experiments use BEVDet as base model with an image size of 256×704 and a ResNet-50 backbone.

\subsubsection{Two Ray-Based Distillation Modules}

Table \ref{table::Ablation studies on distillation modules} validates the effectiveness of our two distillation modules RCD and RWD. The results indicate that each module independently enhances the robustness of the base model, and their combination achieves much higher performance. The RCD module helps the model learn how LiDAR distinguishes accurate and inaccurate locations, while the RWD module minimizes interference from depth-irrelevant information. Notably, two modules can be well applied to all mainstream types of BEV-based models to enhance their robustness.

\subsubsection{Sampling Strategies in RCD}

In the RCD module, we sample negative samples for contrastive learning. We test two sampling strategies: (1) "Random" random samples in background regions along the ray, and (2) "Gaussian" uses a Gaussian distribution to increase sampling probability of regions closer to positive samples. 

Table \ref{table::Ablation studies on sample strategy in RCD} shows that the "Gaussian" strategy achieves better accuracy and robustness, which helps the network more effectively to distinguish nearby positive and negative samples. The results also show that sampling in the teacher network further improves the robustness, because it increases the variety of negative samples.

% % % % % % % % % % % % % % % % % % % % % % % % % % % 

\subsubsection{Calculation of Feature Difference in RWD}

In our weight strategy, we calculate the feature distribution difference between the camera and LiDAR along each ray. We test several strategies. "Cosine similarity" is simple but cannot capture high-dimensional feature differences well. "JS divergence" is less sensitive to small variations between nearby positive and negative samples. Finally, we chose "KL divergence", because it well balances computational cost and performance. Table \ref{table::Ablation studies on feature difference computation strategies in RWD} shows that "KL divergence" achieves the highest accuracy and robustness.

\subsection{Visualization}

Figure \ref{fig:visual} visualizes the detection results of both the representative base model BEVDet and our method on clean data and various types of data corruptions. The results demonstrate that our method consistently detects more objects under all corruption types. However, severe corruption causes information to completely lose in a few regions, leading to lower accuracy even with distillation.

\section{Conclusion}
\label{sec:conclusion}

In this work, we verify that the robustness of multi-view 3D object detection models is limited by inaccurate depth prediction, especially when data corruptions occur in real-world. We first apply the ray prior for cross-modal knowledge distillation and propose a cross-modal distillation method with two novel ray-based modules. It focuses on crucial depth information while mitigating depth-irrelevant information from LiDAR to camera models. Experiments verify the accuracy and robustness of our method on both NuScenes with clean data and RoboBEV with data corruptions. Our method is compatible with all mainstream types of BEV detection models and has the potential to achieve higher accuracy with stronger base models in the future. We will further extend this work to cross-city settings, broader datasets, and more sensor types.

\section{Acknowledgements}

This work was supported by the National Science Foundation of China (NSFC) under Grant 62088102, Fundamental and Interdisciplinary Disciplines Breakthrough Plan of the Ministry of Education of China under Grant JYB2025XDXM504, NSFC under Grants 62373298 and U24A20252, and Nansha Key Science and Technology Project under Grant 2023ZD006.

\bibliography{aaai2026}

\end{document}